# Acute kidney injury prediction for non-critical care patients: a retrospective external and internal validation study

Esra Adiyeke, PhD[a,b‖], Yuanfang Ren, PhD[a,b‖], Benjamin Shickel, PhD[a,b], Matthew M. Ruppert, MS[a,b], Ziyuan Guan, MS[a,b], Sandra L. Kane-Gill, PharmD, MS, FCCM, FCCP[c,d], Raghavan Murugan, MD, MS[d], Nabihah Amatullah, PharmD, RPh[c], Britney A. Stottlemyer, PharmD[c], Tiffany L. Tran, PharmD[c], Dan Ricketts, MET[f], Christopher M. Horvat, MD MHA[d], Parisa Rashidi, PhD[a,e], Azra Bihorac, MD, MS[a,b,*], Tezcan Ozrazgat-Baslanti, PhD[a,b,*]

‖ These first authors have contributed equally.

* These senior authors have contributed equally.

[a] University of Florida Intelligent Clinical Care Center, Gainesville, FL

[b] Department of Medicine, Division of Nephrology, Hypertension, and Renal Transplantation, University of Florida, Gainesville, FL.

[c] Department of Pharmacy and Therapeutics, School of Pharmacy, University of Pittsburgh, Pittsburgh, PA.

[d] Department of Critical Care Medicine, University of Pittsburgh, Pittsburgh, PA.

[e] Department of Biomedical Engineering, University of Florida, Gainesville, FL.

[f] Health Sciences Information Technology, University of Pittsburgh, Pittsburgh, PA.

**Corresponding author:** Azra Bihorac MD, MS, Department of Medicine, Division of Nephrology, Hypertension, and Renal Transplantation, PO Box 100224, Gainesville, FL 32610-0224. Telephone: (352) 294-8580; Fax: (352) 392-5465; Email: abihorac@ufl.edu

**Abstract**

**Background:** Acute kidney injury (AKI), the decline of kidney excretory function, occurs in up to 18% of hospitalized admissions. Progression of AKI may lead to irreversible kidney damage.

**Methods:** This retrospective cohort study includes adult patients admitted to a non-intensive care unit at the University of Pittsburgh Medical Center (UPMC) between July 1, 2018 and December 30, 2022 (n = 46,815) and University of Florida Health (UFH) (n = 127,202) between January 1, 2012 and August 22, 2019. We developed and compared deep learning and conventional machine learning models to predict progression to Stage 2 or higher AKI, defined by Kidney Disease: Improving Global Outcomes serum creatinine criteria, within the next 48 hours. We trained local models for each site (UFH Model trained on UFH, UPMC Model trained on UPMC) and a separate model with a development cohort of patients from both sites (UFH-UPMC Model). We internally and externally validated the models on each site and performed subgroup analyses across sex and race. We presented the features with the highest influence on the model outcomes.

**Results:** The mean age was 55 (standard deviation (SD) 19), 55% (n = 47,350) of the patients were female and 22% (n=18,625) of the patients were African American in the UFH development cohort while the mean age was 71 (SD 14), female patient proportion was 54% (n=15,128) and 11% (n=2,935) of the patients were African-American in UPMC development cohort. Stage 2 or higher AKI occurred in 3% (n=3,257) and 8% (n=2,296) of UFH and UPMC patients, respectively. Area under the receiver operating curve values (AUROC) for the UFH test cohort ranged between 0.77 (UPMC Model) and 0.81 (UFH Model), while AUROC values ranged between 0.79 (UFH Model) and 0.83 (UPMC Model) for the UPMC test cohort. UFH-UPMC Model achieved an AUROC of 0.81 (95% confidence interval [CI] [0.80, 0.83]) for UFH and 0.82 (95% CI [0.81,0.84]) for UPMC test cohorts. Area under the precision recall curve values (AUPRC) for the UFH test cohort was similar for UFH Model (AUPCR, 95% CI, 0.04, [0.04, 0.05]) and UPMC Model (AUPCR, 95% CI, 0.04, [0.04, 0.05]), while AUPRC values ranged between 0.10 (95% CI, [0.09, 0.11]) (UFH Model) and 0.12 (95% CI, [0.10, 0.13]) (UPMC Model) for the UPMC test cohort. UFH-UPMC Model resulted in an AUPRC of 0.6 (95% CI, [0.05, 0.06]) for UFH and 0.13 (95% CI, [0.11,0.15]) for UPMC test cohorts. Kinetic estimated glomerular filtration rate, nephrotoxic drug burden and blood urea nitrogen remained the top three features with the highest influence across the models and health centers.

**Conclusion:** Locally developed models displayed marginally reduced discrimination when tested on another institution, while the top set of influencing features remained the same across the models and sites.

## 1. Introduction

Acute kidney injury (AKI), the decline of kidney excretory function, occurs in up to 18% of acute medical admissions, with the mortality rate amongst these patients averaging 23% and reaching up to 49% for patients who require renal replacement therapy.[1,2] In instances of mild AKI, with little to no kidney cell loss, the patient may achieve full recovery of nephron structure. However, once moderate or severe AKI occurs, the irreversible loss of nephrons may drastically reduce the kidney's lifespan.[2] AKI management with intervention leads to possible early sustained reversal and has been found to yield better prognosis than for patients with late recovery. Thus, timely identification and management of AKI should be implemented to prevent systemic consequences (such as heart failure, liver dysfunction, lung injury, etc.) relating to prolonged imbalanced homeostasis[2,3].

The use of machine-learning techniques with electronic health records (EHR ) has allowed for nonlinear risk calculation to progress positive patient care outcomes via detection.[4-10] Existing AKI risk models have demonstrated limitations that include limited generalizability due to the lack of external validation[5,6], utilizing less representative patient data for model development[9], and focusing on static patient characteristics to predict AKI risk, which failed to capture dynamic changes in risk trajectory over the course of a patient encounter.[7,11] Drug-associated AKI (D-AKI) is a leading cause of AKI accounting for approximately 15-25% of AKI events..[12,13] The negative outcomes of D-AKI are remarkable with up to 70% of patients having residual kidney damage.[14] However, only a few studies have quantified the nephrotoxic burden associated with AKI and it appears that no study has considered it for AKI risk model development.[17]

The objectives of our study are to (1) develop and cross validate (internally and externally validating across health systems) the machine learning model for continuous risk prediction of AKI stage progression (Stage 2 or more) using EHR data for non-critical care patient populations from two hospitals, UFH and UPMC and (2) quantify the nephrotoxic burden and investigate its importance for AKI risk prediction.

## 2. Methods

### *2.1 Study Design and Participants*

A data use agreement (DUA) between the University of Pittsburgh (UPitt) and the University of Florida (UF) was approved for the data transfer needed to validate the proposed algorithm for the Multicenter Implementation of an Electronic Decision Support System for Drug-Associated Acute Kidney Injury (MEnD-AKI) Trial. Under the DUA and following approval by the University of Pittsburgh Institutional Review Board and Pitt Privacy Office (STUDY20120008), UPitt provided de-identified data for 46,815 adult patients who had been admitted to a non-intensive care unit (ICU) at the University of Pittsburgh Medical Center (UPMC) between July 1, 2018 and December 30, 2022. The study was designed and approved by the Institutional Review Board of the University of Florida and the University of Florida Privacy Office (IRB 201901123). Using the University of Florida Health (UFH) Integrated Data Repository as an Honest Broker, we created a single-center, longitudinal dataset extracted from the electronic health records of 127,202 patients ≥18 years admitted to a non-ICU at UFH between January 1, 2012 and August 22, 2019. Each dataset included demographic information, vital signs, laboratory values, medications, and diagnosis and procedure codes for all index admissions. Admissions before index admissions with laboratory values, medications, and diagnoss and procedure codes were included as well.

We identified adult hospitalized patients in non-ICUs with sufficient data excluding (i) patients with end-stage kidney disease (ESKD) or a baseline estimated glomerular filtration rate (eGFR) less than 15 mL/min/1.73m$^2$, (ii) patients who did not have serum creatinine (SCr) within the first two days of hospital admission, (iii) patients who had a hospital length of stay of less than 48 hours. The UFH final cohort included 122,324 hospital encounters from 71,693 patients and the UPMC final cohort included 39,756 patients (Figure 1). We adhered to the guidelines given in Transparent Reporting of a multivariable prediction model for Individual Prognosis or Diagnosis (TRIPOD)[18] and Leisman et al. under the type 3a analysis category[19]. We randomly divided both cohorts into development (70% of observations, n=85,888 encounters for UFH and n=27,808 encounters for UPMC), validation (10%, n=11,879 encounters for UFH and n=3,944 encounters for UPMC), calibration (5%, n=5,981 encounters for UFH and n=1,996 encounters for UPMC) and test (10%, n=18,576 encounters for UFH and n=5,958 encounters for UPMC) sets. All data from each patient was assigned to only one dataset.

*2.2 Outcomes*

Using Kidney Disease: Improving Global Outcomes (KDIGO) guidelines[20] as a foundation, we defined AKI as an increase of SCr ≥0.3 mg/dL within 48 hours, as well as an increase ≥0.3 mg/dL from the baseline, or ≥1.5-fold from the baseline (Figure 2). Baseline SCr was determined using preadmission measurements, or estimated creatinine using the 2021 CKD-EPI refit without race formula (Figure 3).[21,22] AKI severity was defined according to the KDIGO criteria. Stage 2 AKI and Stage 3 AKI with/without renal replacement therapy (RRT) were termed as “moderate to severe AKI”. The outcome of this study was the development of Stage 2 or higher AKI (moderate to severe AKI) within the next 48 hours and was predicted by our model every 12 hours during the entire hospitalization. The AKI stage was calculated when there was a SCr measured and carried forward in time until the next SCr measurement.

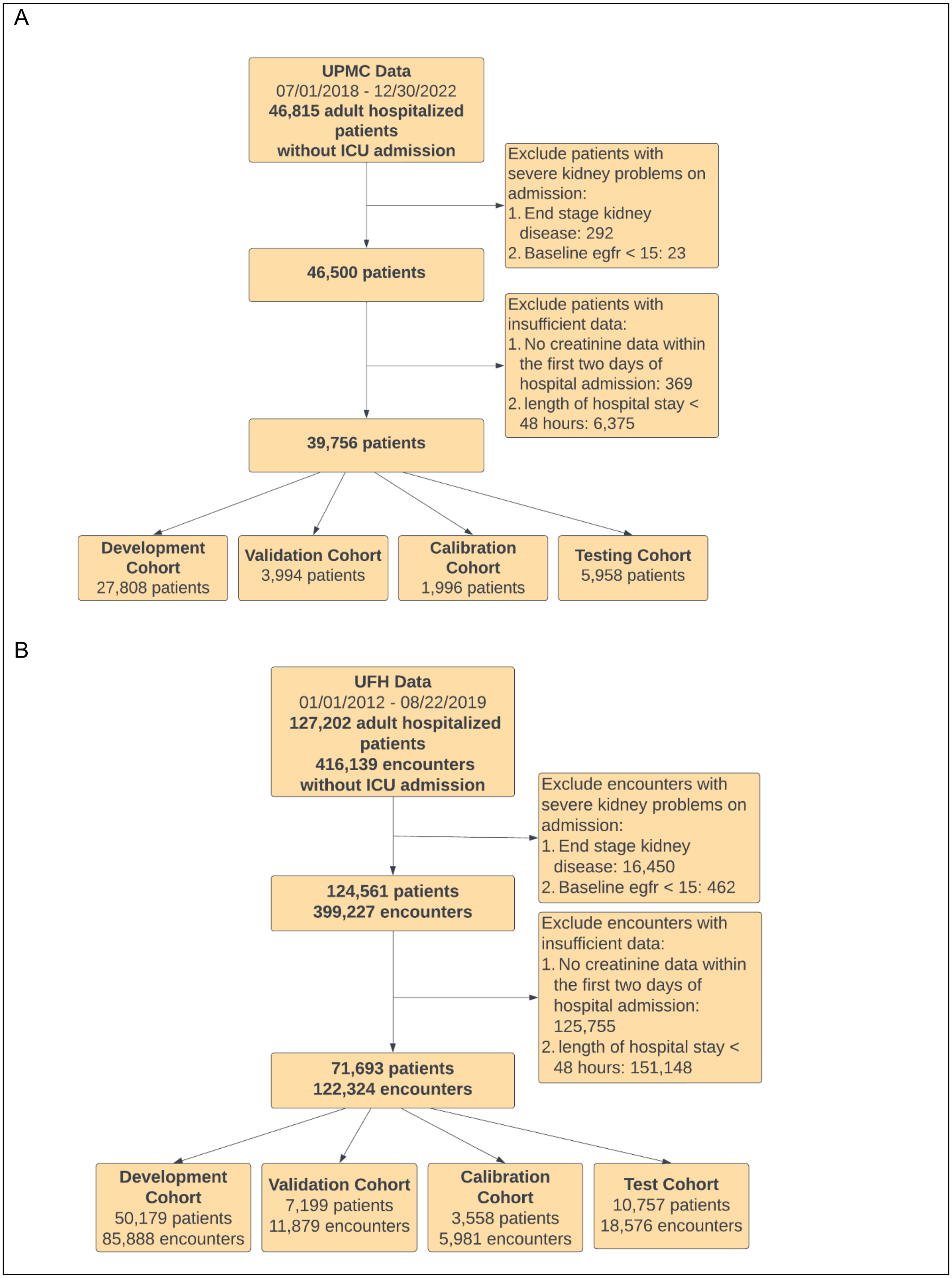

**Figure 1.** Cohort selection and exclusion criteria. A. UFH cohort. B. UPMC cohort.

*2.3. Predictor Variables*

We derived static predictors (features collected on admission) and dynamic predictors (features repeatedly collected during the hospitalization) for both cohorts. Static predictors included demographic features (e.g., age, sex and race), admission features (e.g., admission source), comorbidities, and preadmission medications and laboratory measurements. Supplementary Table 1 provides the list of all features. We derived patient comorbidities using up to 50 International Classification of Disease (ICD) codes from all available historical diagnosis codes. We used validated methods to define binary preadmission comorbidity variables[23,24] and the composite Charlson comorbidity index.[25] We extracted medications dispensed in the one year prior to admission day using RxNorm data grouped into drug classes according to the United States (US) Department of Veterans Affairs National Drug File-Reference Terminology.[26] We extracted laboratory values measured one year prior to admission using Logical Observation Identifiers Names and Codes (LOINC) and calculated derived features, including the count, mean, variance, minimum and maximum values. Missing laboratory variables were imputed using the median values from the training cohort.

Dynamic predictors included kinetic estimated glomerular filtration rate (KeGFR), vital signs (mean arterial pressure (MAP), temperature, respiratory rate and heart rate), laboratory values (basic metabolic panel: sodium, potassium, carbon dioxide, chloride, glucose, calcium and blood urea nitrogen; complete blood count: white blood cell count, hemoglobin, hematocrit, platelets, platelet mean volume, erythrocyte mean corpuscular volume and erythrocyte distribution width; and liver function test: albumin and total bilirubin), the count value of glucose, hemoglobin and albumin representing counts for three panel tests, nephrotoxic medications, and length of stay (Supplementary Table 1). KeGFR was calculated when there was a SCr at least 12 hours apart from the last KeGFR calculation time point. [22]

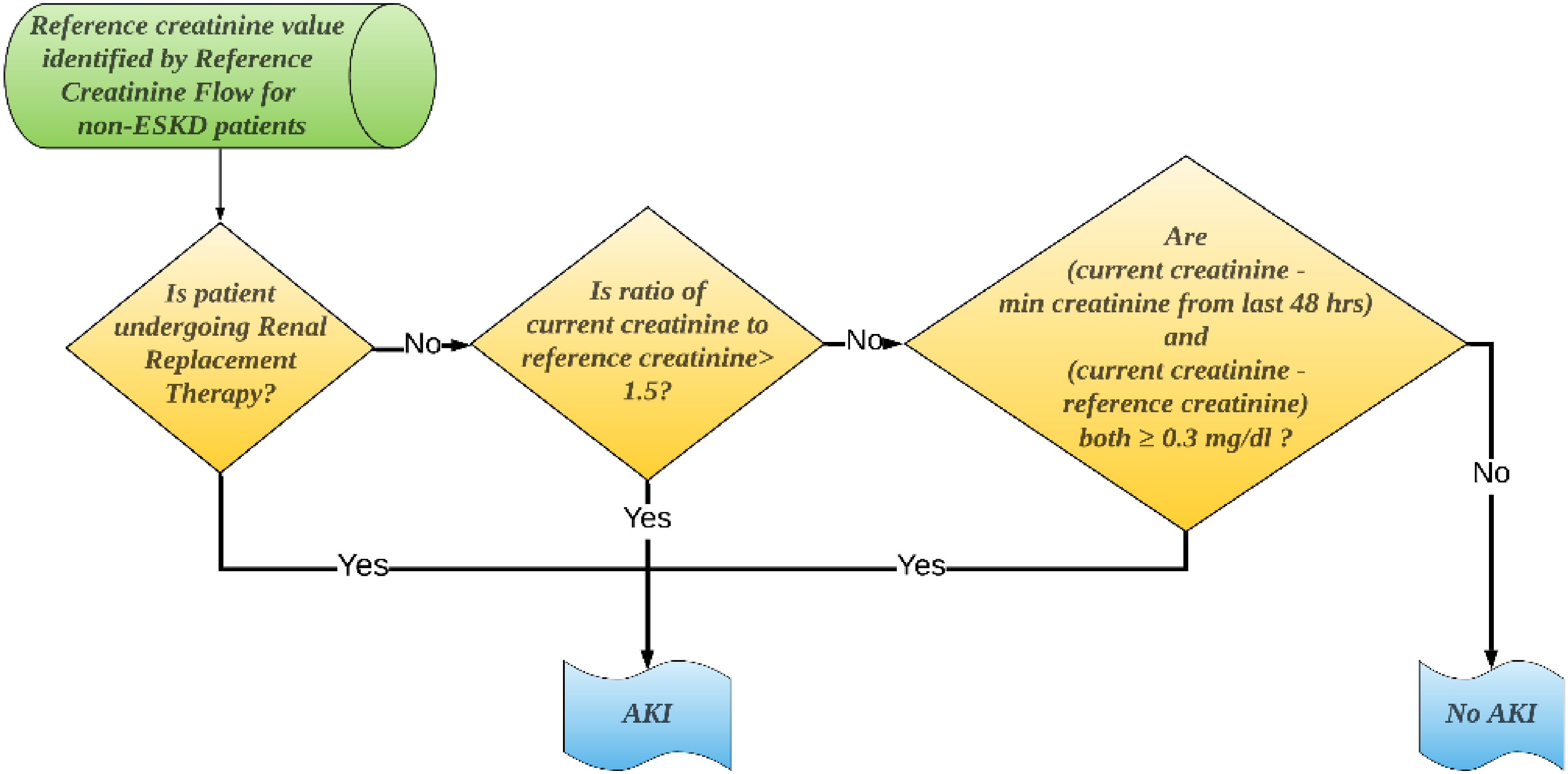

**Figure 2.** AKI identification flow.

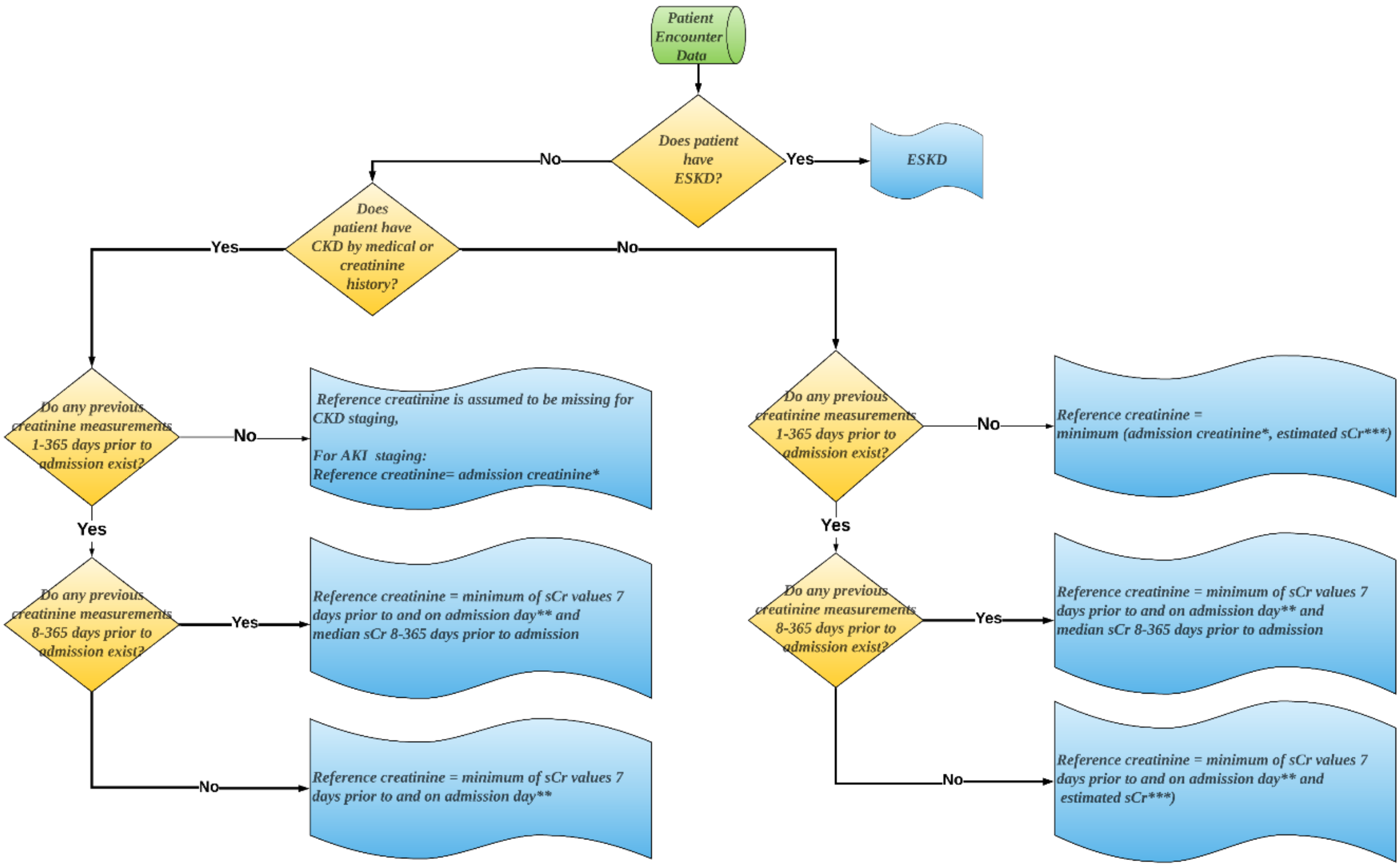

**Figure 3.** Determination of reference creatinine flow. This flow shows the rule for determination reference creatinine changes dynamically during the index admission.

Abbreviations: ESKD, end-stage kidney disease; SCr, serum creatinine.

*Admission creatinine is the first creatinine during encounter.

**Minimum of all SCr values measured from 7 days prior to admission up to first creatinine on admission day is calculated.

***Estimated SCr is obtained by back calculation from the 2021 CKD-EPI refit without race.

Potentially nephrotoxic medications were identified from a previously published modified-Delphi study that assessed the nephrotoxic potential (NxP) of medications used in the non-intensive care setting.[27] Nephrotoxic drugs were assigned scores as 0 for not nephrotoxic, 0.5 for unlikely to possible nephrotoxicity, 1 for possible nephrotoxicity, 1.5 for possible to probable nephrotoxicity, 2 for probable nephrotoxicity, 2.5 for probable to definite nephrotoxicity, and 3 for definite nephrotoxicity. Nephrotoxic drugs that received a NxP score of ≥ 1 (possible nephrotoxicity with routine use) were included. We additionally included combination drug products with a nephrotoxin identified in the modified-Delphi study as an ingredient, along with medications belonging to the same drug class as a nephrotoxin that were not previously included. The nephrotoxin list was updated to include contrast media agents and to add possibly nephrotoxic drugs FDA approved within the past five years as suggested in a review of a drug information database (Lexicomp) and prescribing information. Medications for which nephrotoxicity occurs via acute interstitial nephritis and SCr inhibition (pseudo-AKI) mechanisms were excluded, in addition to the products that are no longer available in the United States and medications administered via the ophthalmic, intraocular, otic, transdermal, topical, and inhalation routes. Additionally, oral vancomycin was excluded. A weight value ranging from 0.2 to 1 was assigned for each nephrotoxin based on the NxP score in the modified-Delphi study, with higher weight values corresponding to higher certainty on nephrotoxicity. For drug combination products, the highest NxP score of the individual drug components was selected to generate the weight value. New nephrotoxins added to the list were arbitrarily assigned a weight value of 0.2.

After extracting static and dynamic predictors, we captured patient status every 12 hours during the entire hospitalization period. For each 12-hour interval, in addition to the static features, we calculated derived features from available dynamic predictors until the end of that interval. Given the different collection frequencies of those dynamic predictors, we derived features using various time windows. That is, we calculated 1) the mean KeGFR within that 12-hour window; 2) the count, mean, variance, minimum and maximum values of each vital sign every 6 hours within the interval; 3) the mean value of each laboratory test and the count value of glucose, hemoglobin and albumin representing three panel tests within the 12-hour interval; 4) the accumulated nephrotoxic burden within the past 7 days by adding the daily nephrotoxic burden (quantified as the sum of weight values for unique daily administrated nephrotoxic drugs); 5) the length of hospital stay from admission to the end of the interval in hours. Missing values of derived features were imputed by carrying forward from the most recent value. If still missing, the median values from the training cohort were used for imputation.

## 3. Model Development and Validation

We trained separate deep learning models on patient populations from UFH (UFH Model), from UPMC (UPMC Model), and a combination of those patient populations from UFH and UPMC (UFH-UPMC Model). Overall structure of the deep learning model consists of a combination of separate subnetworks for processing static nature data, including records collected before admission and patient information revealed throughout their hospitalization. The static sub-network has linear layers for passing and transforming numerical and binarized categorical pre-admission variables. In modeling the longitudinal data, another sub-network based on gated recurrent unit (GRU), that is a sub-class of recurrent neural networks, was used, as this sub-network enabled sequential processing of the data and utilizes dependencies in prior observations. Representations obtained from two sub-networks were concatenated and this unified structure was passed through dense layers, which output the final predicted value.

In training the model, area under the precision-recall curve (AUPRC) was monitored during internal validation. The limit for terminating the model training was defined by three epochs, with no improvement in validation AUPRC. The number of hidden units in the GRU layer was set to 64. A batch size of 64, a learning rate of 1e-3, a drop-out rate of 20% and rectified linear unit (ReLU) activation was used in hidden layers. The model was trained with Adam optimizer, where inversely proportional class weights were used against class imbalance.

To describe feature influences on the predicted outputs, we used integrated gradients (IG) method[28]. By using Shaply Additive Explanations (SHAP) package available in Python, we elucidated the relative contribution of each input variable used in the model[29]. IG method computed the feature attributions by considering each features' effect on the finalized prediction where original feature values and externally introduced independent reference values were used in relevant calculations. Assigned feature attributions were proportional to features' contributions, (i.e., a greater attribution value indicated a higher feature influence). Since the variables were normalized, we used zero as the reference value in calculating feature influences.

We calibrated the model outputs with isotonic regression on hold-out calibration sets derived for each model and assessed model calibration via reliability diagrams given for predicted probabilities and observed outcomes (Supplementary Figure 1). We separately reported the results for independent test cohorts derived for two healthcare systems, i.e. the UFH test set and UPMC test set. The models were trained on all possible 12-hour windows, including steps with starting and trailing Stage 2 and higher AKI windows, while test outcomes were only calculated for the steps for which Stage 2 or higher AKI had not occurred. We evaluated model performance through considering several metrics, including area under the receiver operating characteristic curve (AUROC), AUPRC, accuracy, sensitivity, specificity, positive predictive value (PPV), and negative predictive value (NPV). We tabularized the 95% non-parametric confidence intervals (CI) and calculated for 500 bootstrapped samples by thresholding the probability with the value-optimized Youden index[30]. We performed subgroup analyses and presented AUROC and AUPRC values across the test cohorts stratified by sex (female and male) and race (African American and non-African American). We compared the clinical characteristics and outcomes of patients from two cohorts using the $\chi 2$ test for categorical variables, as well as the Kruskal-Wallis test for continuous variables. All analyses were performed with R 4.3.2 and Python 3.9 software.

## 4. Results

### *4.1 Patient Baseline Characteristics and Outcomes*

Among 85,888 patient encounters in the UFH development cohort, the mean (standard deviation, SD) age was 55 (19) years; 47,350 (55%) encounters were female; 61,589 (72%) encounters were White and 18,625 (22%) were African American; 3,653 (4%) encounters were Hispanic (Table 1). The UPMC development cohort contain 27,808 patient encounters and the demographic characteristics was significantly different from the UFH development cohort (older patients with mean (SD) age, 71 (14) years; 15,128 (54%) females; 24,247 (87%) White and 2,935 (11%) African American; 121 (0%) Hispanic).

Compared with the UFH cohort, patients in the UPMC cohort were more likely to have hypertension (32% vs. 19%) and cardiovascular disease (37% vs. 34%), but less likely to have cancer (15% vs. 26%), liver disease (9% vs. 21%) and chronic kidney disease (25% vs. 27%). Patients in the UFH cohort were more likely to have normal baseline kidney function with significantly higher reference eGFR (97.15 vs. 77.93 mL/min/1.73m$^2$). The median length of hospital stay for both cohorts was 5 days. Compared with the UFH cohort, patients in the UPMC cohort were more likely to develop AKI (36% vs. 14%) and progress to more severe AKI stages (Supplementary Table 2). For patients without AKI, the probability of developing moderate to severe AKI in UPMC and UFH cohorts was 1% and 0.3%, respectively; for patients with Stage 1 AKI, the probability was 11% and 7%, respectively.

**Table 1.** Summary of patient characteristics.

| | UFH development cohort (n=85,888) | UPMC development cohort (n=27,808) | *P*-values |
|---|---|---|---|
| **Demographics** | | | |
| Age, mean (SD), years | 55 (19) | 71 (14) | <0.001 |
| Female sex, n (%) | 47,350 (55) | 15,128 (54) | 0.03 |
| Race, n (%) | | | |
| White | 61,589 (72) | 24,247 (87) | <0.001 |
| African American | 18,625 (22) | 2,935 (11) | <0.001 |
| Other[a] | 4,839 (6) | 412 (1) | <0.001 |
| Missing | 835 (1) | 214 (1) | <0.001 |
| Ethnicity, n (%) | | | |
| Hispanic | 3,653 (4) | 121 (0) | <0.001 |
| Non-Hispanic | 82,235 (96) | 27,687 (100) | <0.001 |
| **Comorbidities** | | | |
| Charlson comorbidity index, median (IQR) | 2 (0, 6) | 2 (0, 5) | <0.001 |
| Hypertension, n (%) | 15,924 (19) | 8,827 (32) | <0.001 |
| Cardiovascular disease, n (%)[b] | 29,402 (34) | 10,236 (37) | <0.001 |
| Diabetes mellitus, n (%) | 24,260 (28) | 7,626 (27) | 0.008 |
| Cancer, n (%) | 22,102 (26) | 4,271 (15) | <0.001 |
| Liver disease, n (%) | 18,254 (21) | 2,596 (9) | <0.001 |
| Chronic kidney disease, n (%) | 23,528 (27) | 6,917 (25) | <0.001 |
| Estimated glomerular filtration rate, median (IQR), mL/min/1.73m$^2$ | 97.15 (78.47, 111.81) | 77.93 (54.14, 94.90) | <0.001 |
| **Outcome** | | | |
| Hospital length of stay, median (IQR), days | 5 (3, 7) | 5 (4, 8) | <0.001 |
| Worst AKI stage, n (%) | | | |
| No AKI | 73,780 (86) | 17,724 (64) | <0.001 |
| Stage 1 | 8,851 (10) | 7,788 (28) | <0.001 |
| Stage 2 | 2,003 (2) | 1,806 (6) | <0.001 |
| Stage 3 | 1,166 (1) | 438 (2) | 0.008 |
| Stage 3 + RRT | 88 (0.1) | 52 (0.2) | <0.001 |

AKI: acute kidney injury; RRT: renal replacement therapy; SD: standard deviation; IQR: interquartile range.
[a] Other race includes American Indian/Alaskan Native, Asian, Native Hawaiian/other Pacific Islander, and multiracial.
[b] Cardiovascular disease was considered if there was a history of congestive heart failure, coronary artery disease of peripheral vascular disease.

### *4.2 Model Performance*

We reported trained model performance outcomes for predicting Stage 2 or higher AKI within the next 48 hours on hold-out test cohorts obtained for each health system (UFH and UPMC) separately and presented AUROC, AUOPRC, sensitivity, specificity, PPV and NPV results with their 95% CI in Table 2. Both UFH Model (AUROC and 95% CI on UFH test cohort 0.81 [0.79, 0.82] vs UPMC test cohort 0.79 [0.78, 0.8]) and UPMC Model (AUROC, 95% CI UPMC test cohort 0.82 [0.81, 0.83] vs 0.77 [0.75, 0.78]) had higher AUROC values for their source sites, while the models had reduced AUROC values for the test patient populations derived from different from health systems. For the UFH test patient population, AUPRC values were similar for UFH Model (AUPRC 0.05 [95% CI 0.04, 0.06]) and UPMC Model (0.05 [0.04, 0.06]), yet slightly higher for UFH-UPMC Model (0.06 [0.05, 0.06]). While the best AUPRC values on UPMC test cohort were obtained for Model 3 (0.13 [0.11, 0.15]), UFH Model output the lowest value (0.1 [0.09, 0.11]). Computed AUPRC values on the UFH test cohort were significantly lower than the UPMC for all three models. Sensitivity was highest for the models performing on their source site (Sensitivity, 95% CI for UFH Model on UFH 0.81 [0.81, 0.81] and for UPMC Model on UPMC 0.82 [0.82, 0.82]), while specificity was the highest for UFH-UPMC Model (Specificity, 95% CI for UFH-UPMC Model on UFH 0.72 [0.7, 0.75] and on UPMC 0.73 [0.71, 0.76]).

Model discrimination for patient populations across the sites with respect to their demographic profiles were evaluated with AUROC and AUPRC (Table 3). While all three models had slightly higher AUROC values for the female subgroup in the UFH health system, this trend was reversed in the UPMC test cohort, as male patients had better AUROC outcomes than female patients. Similarly, models resulted in better AUPRC values for female subjects than males for the UFH test cohort and higher AUPRC outcomes for male patients than females for the UPMC test cohort. While not statistically significant, UFH Model had higher AUROC and AUPRC for non-African American patients (AUROC, 95% CI for non-African American patients for the UFH 0.81 [0.80, 0.82] and for the UPMC 0.79 [0.78, 0.8]) than African Americans on both sites (AUROC, 95% CI for African American patients on UFH 0.79 [0.76, 0.81] and on UPMC 0.78 [0.73, 0.81]).

**Table 2.** Classification performance metrics for each model on test cohorts from separate health institutions.

| | Sensitivity (95% CI) | Specificity (95% CI) | PPV (95% CI) | NPV (95% CI) | Accuracy (95% CI) | AUROC (95% CI) | AUPRC (95% CI) |
|---|---|---|---|---|---|---|---|
| **UFH Model** | | | | | | | |
| UFH test cohort | 0.81<br>(0.81, 0.81) | 0.66<br>(0.63, 0.68) | 0.02<br>(0.02, 0.02) | 1.0<br>(1.0, 1.0) | 0.81<br>(0.81, 0.81) | 0.81<br>(0.79, 0.82) | 0.04<br>(0.04, 0.05) |
| UPMC test cohort | 0.78<br>(0.77, 0.78) | 0.64<br>(0.62, 0.67) | 0.06<br>(0.05, 0.06) | 0.99<br>(0.99, 0.99) | 0.77<br>(0.77, 0.78) | 0.79<br>(0.78, 0.8) | 0.1<br>(0.09, 0.11) |
| **UPMC Model** | | | | | | | |
| UFH test cohort | 0.76<br>(0.76, 0.76) | 0.63<br>(0.61, 0.66) | 0.02<br>(0.02, 0.02) | 1.0<br>(1.0, 1.0) | 0.76<br>(0.76, 0.76) | 0.77<br>(0.75, 0.78) | 0.04<br>(0.04, 0.05) |
| UPMC test cohort | 0.82<br>(0.82, 0.82) | 0.69<br>(0.66, 0.71) | 0.07<br>(0.07, 0.08) | 0.99<br>(0.99, 0.99) | 0.82<br>(0.82, 0.82) | 0.83<br>(0.82, 0.84) | 0.12<br>(0.1, 0.13) |
| **UFH-UPMC Model** | | | | | | | |
| UFH test cohort | 0.77<br>(0.77, 0.77) | 0.72<br>(0.7, 0.75) | 0.02<br>(0.02, 0.02) | 1.0<br>(1.0, 1.0) | 0.77<br>(0.77, 0.77) | 0.81<br>(0.8, 0.83) | 0.06<br>(0.05, 0.06) |
| UPMC test cohort | 0.76<br>(0.75, 0.76) | 0.73<br>(0.71, 0.76) | 0.06<br>(0.06, 0.06) | 0.99<br>(0.99, 0.99) | 0.76<br>(0.75, 0.76) | 0.82<br>(0.81, 0.84) | 0.13<br>(0.11, 0.15) |

CI: confidence interval; PPV: positive predictive value; NPV: negative predictive value; AUPRC: area under the precision-recall curve; AUROC: area under the receiver operating characteristic curve; RNN: recurrent neural network; MLP: multi-layer perceptron.
UFH Model was trained on UF development dataset, UPMC Model was trained on UPMC development dataset and UFH-UPMC Model was trained on combination of UFH and UPMC development datasets.

**Table 3.** Classification performance metrics for each model on test cohorts from separate health institutions stratified by sex and race.

| | All | | Female | | Male | | African American | | Non-African American | |
|---|---|---|---|---|---|---|---|---|---|---|
| | **AUROC (95% CI)** | **AUPRC (95% CI)** | **AUROC (95% CI)** | **AUPRC (95% CI)** | **AUROC (95% CI)** | **AUPRC (95% CI)** | **AUROC (95% CI)** | **AUPRC (95% CI)** | **AUROC (95% CI)** | **AUPRC (95% CI)** |
| **UFH Model** | | | | | | | | | | |
| UFH | 0.81 (0.79, 0.82) | 0.04 (0.04, 0.05) | 0.81 (0.79, 0.83) | 0.05 (0.04, 0.05) | 0.8 (0.78, 0.82) | 0.04 (0.04, 0.05) | 0.79 (0.76, 0.81) | 0.04 (0.04, 0.06) | 0.81 (0.8, 0.82) | 0.04 (0.04, 0.05) |
| UPMC | 0.79 (0.78, 0.8) | 0.1 (0.09, 0.11) | 0.78 (0.77, 0.8) | 0.09 (0.08, 0.11) | 0.8 (0.78, 0.81) | 0.1 (0.09, 0.12) | 0.78 (0.73, 0.81) | 0.08 (0.06, 0.11) | 0.79 (0.78, 0.8) | 0.1 (0.09, 0.11) |
| **UPMC Model** | | | | | | | | | | |
| UFH | 0.77 (0.75, 0.78) | 0.04 (0.04, 0.05) | 0.78 (0.76, 0.79) | 0.04 (0.04, 0.05) | 0.75 (0.73, 0.78) | 0.04 (0.03, 0.05) | 0.72 (0.69, 0.75) | 0.03 (0.02, 0.04) | 0.78 (0.76, 0.8) | 0.05 (0.04, 0.06) |
| UPMC | 0.83 (0.82, 0.84) | 0.12 (0.1, 0.13) | 0.81 (0.79, 0.82) | 0.1 (0.09, 0.12) | 0.86 (0.84, 0.87) | 0.14 (0.12, 0.16) | 0.84 (0.81, 0.87) | 0.1 (0.07, 0.13) | 0.83 (0.82, 0.84) | 0.12 (0.11, 0.14) |
| **UFH-UPMC Model** | | | | | | | | | | |
| UFH | 0.81 (0.8, 0.83) | 0.06 (0.05, 0.06) | 0.83 (0.81, 0.84) | 0.06 (0.05, 0.08) | 0.8 (0.78, 0.82) | 0.05 (0.04, 0.06) | 0.8 (0.78, 0.83) | 0.05 (0.04, 0.06) | 0.82 (0.8, 0.83) | 0.06 (0.05, 0.07) |
| UPMC | 0.82 (0.81, 0.84) | 0.13 (0.11, 0.15) | 0.82 (0.8, 0.83) | 0.11 (0.1, 0.13) | 0.83 (0.81, 0.85) | 0.15 (0.12, 0.18) | 0.85 (0.81, 0.87) | 0.12 (0.09, 0.16) | 0.82 (0.81, 0.83) | 0.13 (0.11, 0.15) |

CI: confidence interval; PPV: positive predictive value; NPV: negative predictive value; AUPRC: area under the precision-recall curve; AUROC: area under the receiver operating characteristic curve; RNN: recurrent neural network; MLP: multi-layer perceptron.
UFH Model; was trained on UF development dataset, UPMC Model was trained on UPMC development dataset and UFH-UPMC Model was trained on combination of UFH and UPMC development datasets.

UPMC Model resulted in higher AUROC values on UFH for non-African Americans (AUROC, 95% CI for non-African-Americans 0.78 [0.76, 0.8] vs for African Americans 0.72 [0.69, 0.75]), while those outcomes were better for African American patients on UPMC sites (AUROC, 95% CI for African Americans 0.84 [0.81, 0.87] vs for non-African Americans 0.83 [0.82, 0.84]). Similarly, UFH-UPMC Model had higher AUROC an AUPRC values for non-African Americans on UFH and for African Americans on UPMC patient populations.

*4.3 Comparison to baseline models*

We compared our developed model to three baseline models, logistic regression, random forest, and extreme gradient boosting (XGBoost) models (Table 4). Like our developed model, baseline models made predictions every 12 hours during hospitalization. For each prediction model, fixed static predictors and dynamic predictors from the preceding 48 hours were used. Thus, predictions were made only 48 hours after hospital admission. XGBoost model achieved the best performance with sensitivity 0.82 (95% CI, 0.81-0.82), specificity 0.78 (95% CI, 0.76-0.8), AUROC 0.87 (95% CI, 0.86-0.88) and AUPRC 0.1 (95% CI, 0.09-0.11) on test cohort dataset from both cohorts. UFH-UPMC Model had the second-best performance with sensitivity 0.77 (95% CI, 0.77-0.78), specificity 0.78 (95% CI, 0.76-0.8), AUROC 0.85 (95% CI, 0.84-0.86) and AUPRC 0.09 (95% CI, 0.08-0.1). Though the XGBoost model outperformed our model, differences between two models' performance were not statistically significant in terms of AUROC, AUPRC and specificity and our model had the ability to predict early AKI outcomes (first 48 hours within hospital admission).

**Table 4.** Comparison of different predictive models.

| Model Name | Sensitivity (95% CI) | Specificity (95% CI) | PPV (95% CI) | NPV (95% CI) | Accuracy (95% CI) | AUROC (95% CI) | AUPRC (95% CI) |
|---|---|---|---|---|---|---|---|
| Logistic Regression | 0.8 (0.79, 0.8) | 0.73 (0.71, 0.75) | 0.03 (0.03, 0.04) | 1.0 (1.0, 1.0) | 0.8 (0.79, 0.8) | 0.83 (0.82, 0.84) | 0.07 (0.06, 0.08) |
| Random Forest | 0.83 (0.83, 0.83) | 0.73 (0.71, 0.75) | 0.04 (0.04, 0.04) | 1.0 (1.0, 1.0) | 0.83 (0.83, 0.83) | 0.84 (0.83, 0.85) | 0.07 (0.07, 0.08) |
| XGBoost | 0.82 (0.81, 0.82) | 0.78 (0.76, 0.8) | 0.04 (0.04, 0.04) | 1.0 (1.0, 1.0) | 0.81 (0.81, 0.82) | 0.87 (0.86, 0.88) | 0.1 (0.09, 0.11) |
| RNN based model (UFH-UPMC Model) | 0.77 (0.77, 0.78) | 0.78 (0.76, 0.8) | 0.03 (0.03, 0.03) | 1.0 (1.0, 1.0) | 0.77 (0.77, 0.78) | 0.85 (0.84, 0.86) | 0.09 (0.08, 0.1) |

XGBoost: extreme gradient boosting; RNN: recurrent neural network.
All models were trained using development cohort dataset from both cohorts. Performance was evaluated on test cohort dataset from both cohorts.
As logistic regression, random forest and XGBoost models used preceding 48 hours data for predictions, predictions were made only after 48 hours after hospital admission.

*4.4 Interpreting Model Outputs*

We presented the list of top 20 input variables with the highest feature importance values for the models predicting Stage 2 or higher AKI within the next 48 hours. We illustrated the mean absolute SHAP values for aggregated test patient populations from UFH (Figure 4 A, Figure 5A), from UPMC (Figure 4B, Figure 5B) and the combination of test cohorts from both institutions (Figure 6). The order of the three most important variables was consistent across sites (both in terms of source and target sites) and models. Those variables were mean KeGFR in the last 12 hours, nephrotoxic drug burden that occurred in the last 7 days and mean blood urea nitrogen in the last 12 hours, respectively. Noting that, SHAP values do not constitute or indicate any causal relation between features and outcome(s). Thus, those top three variables could be considered as plausible variables that potentially signal patients' future kidney conditions, as they were shown to be biologically correlated to kidney damage or recovery.[12,31,32]

When examined separately, we observed that the majority of the top 10 variables were based on routine laboratory measurements repeatedly collected during patient hospitalizations for UFH Model and UPMC Model on both the UFH and the UPMC health systems. Specifically, mean serum chloride in the past 12 hours and mean serum sodium in the past 12 hours were consistently listed within the top 10 important features derived for UFH Model and UPMC Model for both health systems. However, for UFH-UPMC Model, we observed mean serum calcium in the past 12 hours. Despite not necessarily being direct indicators of the patients' most recent kidney conditions, SCr variables derived within 365 days before admission were clinically intuitive, as those values were potentially used to determine patients' baseline SCr values. In line with this, minimum SCr values within the past year before admission consistently remained among the top influential features across the models and sites.

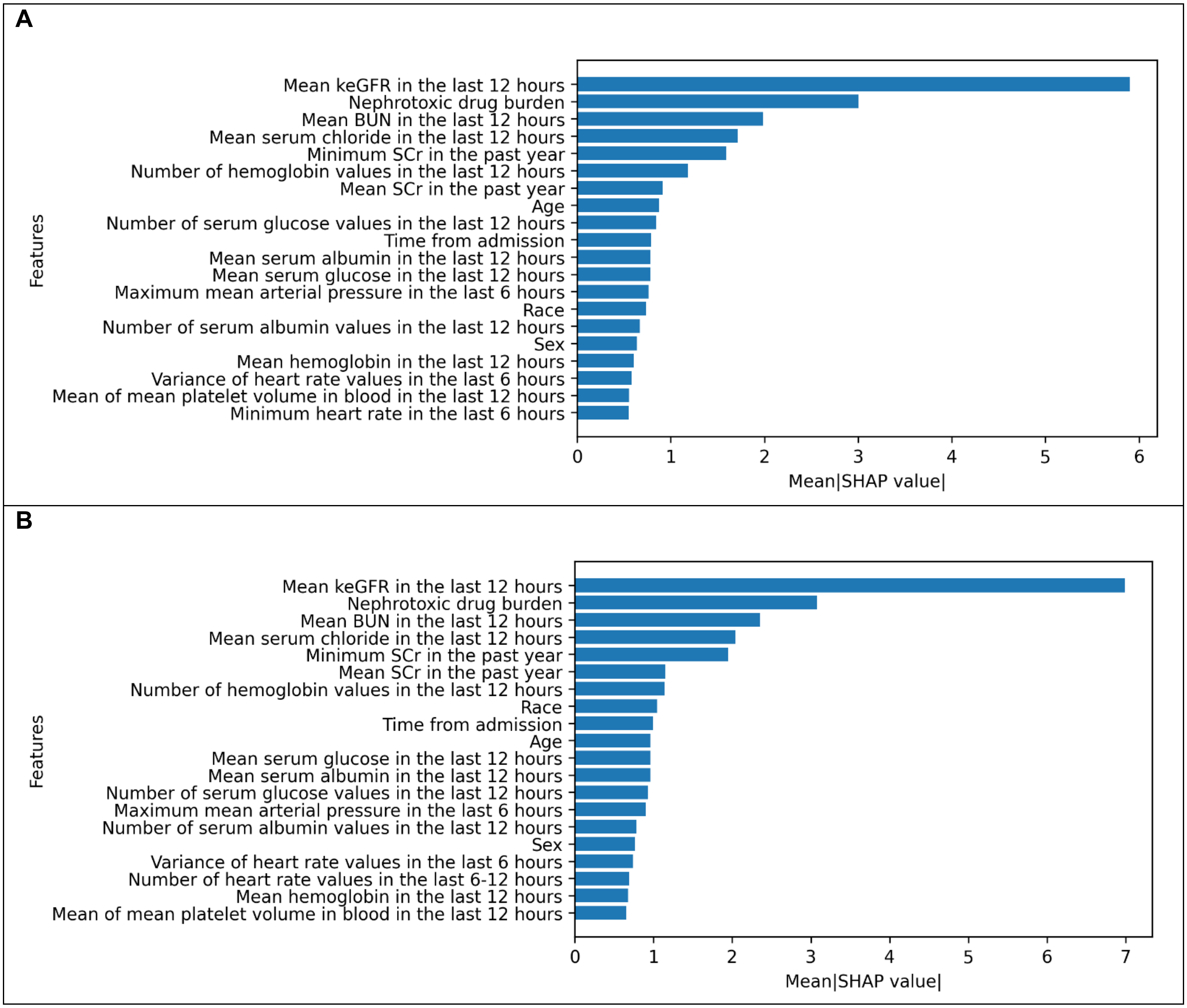

**Figure 4.** Feature importance for UFH Model on UFH test set (A) and on UPMC test set (B).

Vital records (nadir value of body temperature in the past 6-12 hours and the maximum mean arterial pressure in the last 6 hours) were indexed among the top 10 variables for UFH-UPMC Model for the combined test population. Other common variables observed among the top 20 features across all models and health systems were based on the count variables of available laboratory values monitoring metabolic (number of serum albumin values in the past 12 hours) and blood (number of hemoglobin values in the past 12 hours) indicators. Patients' demographic variables, specifically age and sex, had higher influence on predictions made with UFH Model and UPMC Model on both sites, while race was also listed among the top 20 for UFH Model on both the UFH and the UPMC centers. Apart from minimum SCr value, all variables with the highest influence on the predictions made with UFH-UPMC Model were derived from records streamed during hospitalization.

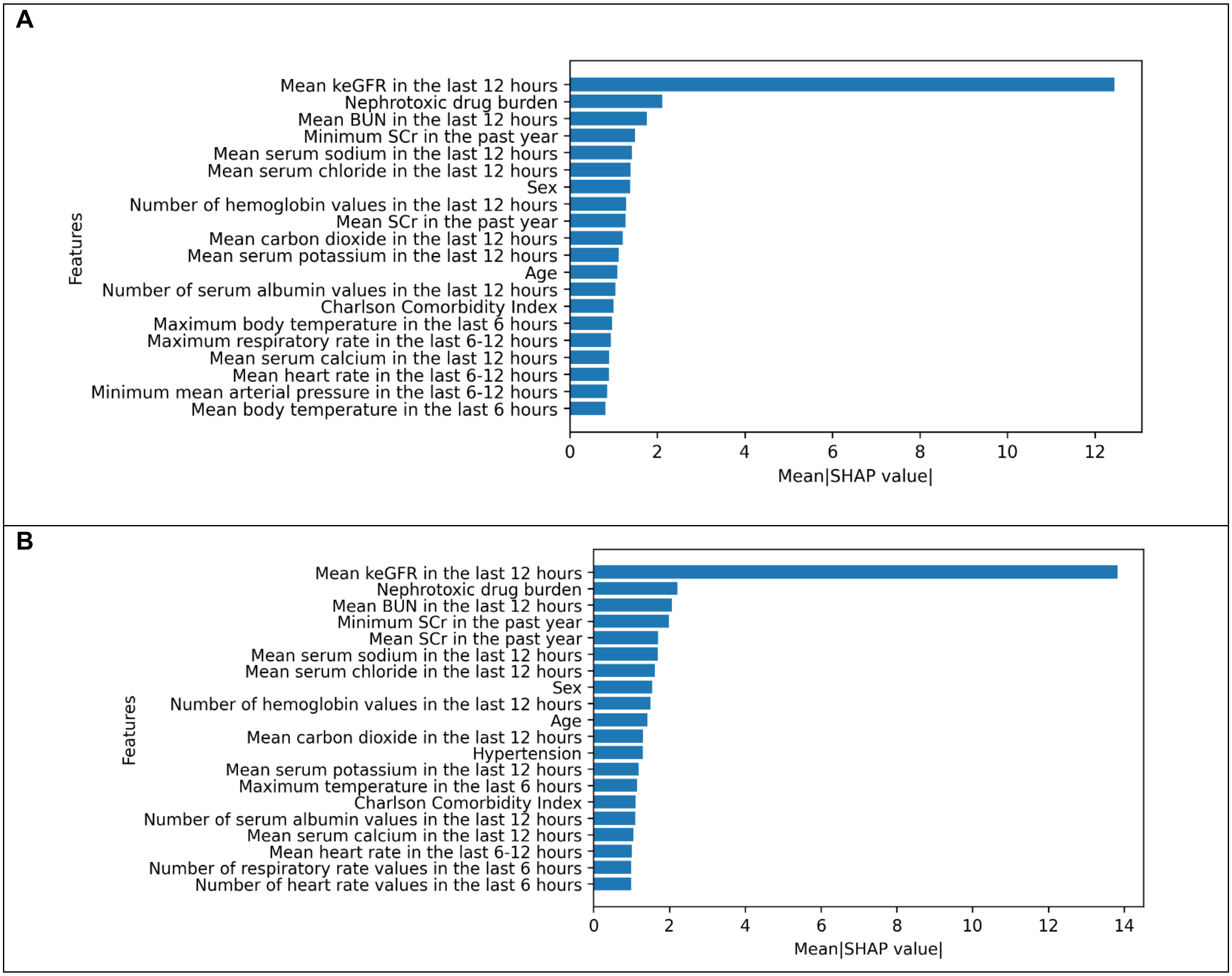

**Figure 5.** Feature importance for UPMC Model on UFH test set (A) and on UPMC test set (B).

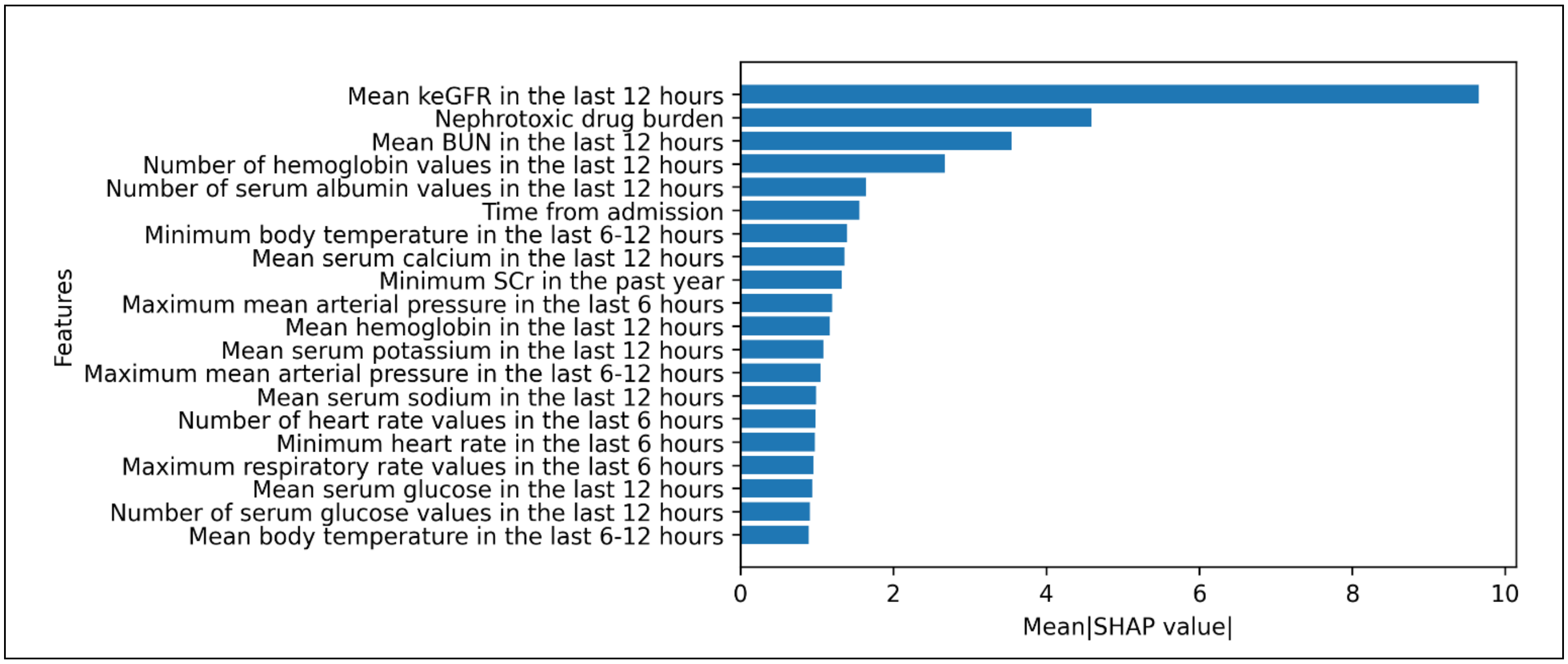

**Figure 6.** Feature importance for UFH-UPMC Model on combined test sets from UFH and UPMC.

## 5. Discussions

In this study, we developed models for continuously predicting Stage 2 or higher AKI within the next 48 hours for patient populations drawn from UFH and UPMC, which are two geographically separated and diversely different health centers operating on different integrated information and data warehousing systems. We harmonized the de-identified data obtained from those two health systems to eliminate health record heterogeneities and ensure data format consistency and interoperability within the established model framework. Built models were operated on 12-hour time window basis by leveraging a rich set of input variables obtained from EHR data, e.g. demographics, laboratory and vital measurements, diagnoses, and medications. We trained separate models for cohorts derived from each health system only and tested the models' performance on both source and the external health systems. As a second strategy for improving diversity in the training data set, we developed models on a larger training set by combining the patients from those health systems and evaluated the outcomes for UFH and UPMC. We performed subgroup analyses across patient demographics — including sex and race — and reported performance outcomes for each health system separately. We elucidated feature importance by using interpretability methods suitable for the developed models.

Recent reviews have demonstrated that AKI is an active field of research for improved predictive, diagnostic, and prognostic digital health solutions[33,34]. In line with that, Yu et al. described the expanded literature on machine learning based studies for predicting AKI for all-care, critical care and surgical patients, in addition to the applications tailored for specific conditions, e.g. sepsis, burn, and nephrotoxic exposures[35]. Unlike past reviews focusing on sub-specialized patient cohorts[36], care levels[11], modeling approaches[37] or modeling objectives, Feng et al. have reviewed AKI risk prediction models from a broader perspective by considering patients populations from all clinical settings and underlined the importance of standardized outcome definitions, as well as consistency across the predictive variables for improved external validation[38]. In addition, difficulties in external validation of AKI prediction models developed for general or non-critical care patients due to lack of publicly available data were specifically highlighted in another recent review given by Uchino et al.[39]

While most AKI prediction models were targeting critical care populations, several studies focused on general patient populations admitted to hospitals for receiving non-critical care. Notably, in a systemic review given by Wainstein et al. only three machine learning studies were reported for predicting AKI where used AKI definition conforms with recent KDIGO criteria and externally validated on adult patient populations from the same institution with temporal difference or from a distinct health system or centers.[40] Churpek and colleagues considered patients admitted to three health systems in the United States and validated a gradient boosted machine (GBM) based model for continuously predicting Stage 2 or higher AKI within the next 48 hours with reported AUROC values of 0.80 and 0.84 for ward patients in the external validation cohorts.[41] Similarly, US based patients who were admitted to six sites in five states were considered for training and validating models for predicting Stage 2 or higher AKI within the next 48 hours in Song et al.[42] In this study, authors performed external validation experiments on five different sites for the model trained on the selected source site and reported AUROC values between 0.68 and 0.80. General patients admitted to two tertiary hospitals in Korea were considered in Kim et al. where AKI within the next 7 days was predicted with an RNN based model.[43] Their model achieved an AUROC value for 0.90 for stage 2 or higher AKI in their external validation experiments. In addition to the aforementioned three articles, we believe that it would be worth including the recent work done by Cao and colleagues, in which the authors replicated a state-of-the-art AKI prediction study given by Tomasev et al. on a different and more diverse patient population and examined the model performance.[9,10] The model given in Cao et al. is not an exact replication, but an approximation to the model presented in Tomasev et al. due to access and processing capacity issues related with source data and data platform. Cao and colleagues reported an AUROC value of 0.81 for the model predicting Stage 2 or higher AKI within the next 48 hours and performed further experiments for sex and race.

In our study, we used a deep learning model that dynamically predicts stage 2 or higher AKI within the next 48 hours in line with evolving landscape of AKI prediction models as discussed in prior studies.[44,45] Accurately predicting the patients who are likely to get worse with a reasonable lead time, may enable care-givers to capture the cases that need more aggressive treatments, changing the fluid administration plan or re-evaluate and control the nephrotoxic exposure.[34,42] Our study departs from Churpek et al. by providing sex and race subgroup analyses, which allows us to evaluate performance discrepancies across these variables. Our work also differs from the external validation study provided by Cao et al. as we consider the complete hospitalization period of patients in models while Cao and colleagues only modeled the first 7 days of admission. Similarly, in Song et al. patients' length of stays were limited to the first 7 days of admissions, therefore time periods beyond 7 days remained unexamined. Our outcome definition is based on predicting Stage 2 and higher AKI, while in Kim et al., outcome was based on AKI within 7 days. Our study contributes to AKI prediction research by developing a deep dynamic model for non-critical care patients while maintaining competitive performance results for externally validated patient populations.

Our study has several limitations. First, the AKI definition used in this study was based on KDIGO SCr criteria rather than urine output criteria. This was because urine output measurements taken during general care clinics visits remain unreliable and are not monitored as intensively as in critical care units. Second, although we aimed to maintain a clinically relevant and rich set of features in our model pipeline, we were limited with respect to the data elements incorporated in the health systems, since all elements used in the model could be generated for both UFH and UPMC patient populations. This resulted in eliminating some potentially pertinent variables, as they were unavailable for either site. Third, we relied on patients' electronic health records to derive nephrotoxic drug related features and did not consider over the counter nephrotoxins or home medications.  Potential missing undocumented nephrotoxic drugs, e.g., non-steroidal anti-inflammatory drugs, could be identified to some extent via natural language processing techniques applied on clinical notes. Fourth, the model we used in this study was based on recurrent neural network architecture, which is known to have limitations processing long sequences due to vanishing gradients. Therefore, the model may fail to deliver accurate results for patients with long hospitalization period. Fifth, we used a retrospective study design which does not have delays in accessing and processing data, hence, it requires prospective validation for real-time clinical utility.

## 6. Conclusions

In this work, we developed externally validated models for predicting Stage 2 and higher AKI within the next 48 hours for patients admitted to general hospital for two separate health systems. Our model maintained competitive performance on external validation patient populations. We performed subgroup analyses across demographics for both sites and reported the feature importance for deep models on sites in which the rankings were in line with clinical intuition in identifying AKI outcome. Although externally validated models and results presented in this paper were strengths of this study, real-time clinical beneficiaries of the model necessitate further evaluation, which is an immediate objective of our future work.

**Acknowledgments:** We gratefully acknowledge the technical support of NVIDIA AI Technology Center (NVAITC) at UF for this research. We would like to acknowledge the Intelligent Clinical Care Center research group for support provided for this study. We acknowledge the University of Florida Integrated Data Repository (IDR) and the UF Health Office of the Chief Data Officer for providing the analytic data set for this project.

**Author contributions:** EA, YR, BS, RM, SKG, PR, AB, TOB, BAS, TT and NA contributed to the study design. EA, YR, and TOB drafted the manuscript. EA, YR, ZG, MR, DR, CH worked on data processing and analysis. All authors contributed to data interpretation and provided critical revisions.

**Funding**

A.B., T.O.B., S.K.G., R.M., P.R., Y.R., and B.S. were supported by R01 DK121730, and E.A. and T.O.B. were supported by K01 DK120784 from the National Institute of Diabetes and Digestive and Kidney Diseases (NIH/NIDDK). This work was also supported in part by the NIH/NCATS Clinical and Translational Sciences Award to the University of Florida UL1 TR000064. The content is solely the responsibility of the authors and does not necessarily represent the official views of the National Institutes of Health. The funders had no role in design and conduct of the study; collection, management, analysis, and interpretation of the data; preparation, review, or approval of the manuscript; and decision to submit the manuscript for publication. The authors declare that they have no conflict of interests. A.B. and T.O.B. had full access to all the data in the study and take responsibility for the integrity of the data and the accuracy of the data analysis. E.A. and Y.R. conducted and are responsible for the data analysis.

## References

1. Finlay S, Bray B, Lewington AJ, et al. Identification of risk factors associated with acute kidney injury in patients admitted to acute medical units. *Clin Med (Lond)* 2013; 13(3): 233-8.
2. Kellum JA, Romagnani P, Ashuntantang G, Ronco C, Zarbock A, Anders H-J. Acute kidney injury. *Nature Reviews Disease Primers* 2021; 7(1): 52.
3. Kellum JA, Sileanu FE, Bihorac A, Hoste EA, Chawla LS. Recovery after Acute Kidney Injury. *Am J Respir Crit Care Med* 2017; 195(6): 784-91.
4. Koyner JL, Adhikari R, Edelson DP, Churpek MM. Development of a Multicenter Ward-Based AKI Prediction Model. *Clin J Am Soc Nephrol* 2016; 11(11): 1935-43.
5. Koyner JL, Carey KA, Edelson DP, Churpek MM. The Development of a Machine Learning Inpatient Acute Kidney Injury Prediction Model. *Crit Care Med* 2018; 46(7): 1070-7.
6. Haines RW, Lin SP, Hewson R, et al. Acute Kidney Injury in Trauma Patients Admitted to Critical Care: Development and Validation of a Diagnostic Prediction Model. *Sci Rep* 2018; 8(1): 3665.
7. Peng JC, Wu T, Wu X, et al. Development of mortality prediction model in the elderly hospitalized AKI patients. *Sci Rep* 2021; 11(1): 15157.
8. Motwani SS, McMahon GM, Humphreys BD, Partridge AH, Waikar SS, Curhan GC. Development and Validation of a Risk Prediction Model for Acute Kidney Injury After the First Course of Cisplatin. *J Clin Oncol* 2018; 36(7): 682-8.
9. Tomašev N, Glorot X, Rae JW, et al. A clinically applicable approach to continuous prediction of future acute kidney injury. *Nature* 2019; 572(7767): 116-9.
10. Cao J, Zhang X, Shahinian V, et al. Generalizability of an acute kidney injury prediction model across health systems. *Nat Mach Intell* 2022; 4(12): 1121-9.
11. Hodgson LE, Sarnowski A, Roderick PJ, Dimitrov BD, Venn RM, Forni LG. Systematic review of prognostic prediction models for acute kidney injury (AKI) in general hospital populations. *BMJ Open* 2017; 7(9): e016591.
12. Mehta RL, Pascual MT, Soroko S, et al. Spectrum of acute renal failure in the intensive care unit: the PICARD experience. *Kidney Int* 2004; 66(4): 1613-21.
13. Uchino S, Kellum JA, Bellomo R, et al. Acute renal failure in critically ill patients: a multinational, multicenter study. *JAMA* 2005; 294(7): 813-8.
14. Menon S, Kirkendall ES, Nguyen H, Goldstein SL. Acute kidney injury associated with high nephrotoxic medication exposure leads to chronic kidney disease after 6 months. *J Pediatr* 2014; 165(3): 522-7 e2.
15. Davison AM, Jones CH. Acute interstitial nephritis in the elderly: a report from the UK MRC Glomerulonephritis Register and a review of the literature. *Nephrol Dial Transplant* 1998; 13 Suppl 7: 12-6.
16. Baker RJ, Pusey CD. The changing profile of acute tubulointerstitial nephritis. *Nephrol Dial Transplant* 2004; 19(1): 8-11.
17. Adiyeke E, Ren Y, Ruppert MM, et al. A deep learning-based dynamic model for predicting acute kidney injury risk severity in postoperative patients. *Surgery* 2023; 174(3): 709-14.
18. Collins GS, Reitsma JB, Altman DG, Moons KG. Transparent reporting of a multivariable prediction model for individual prognosis or diagnosis (TRIPOD): the TRIPOD Statement. *BMC Med* 2015; 13: 1.
19. Leisman DE, Harhay MO, Lederer DJ, et al. Development and Reporting of Prediction Models: Guidance for Authors From Editors of Respiratory, Sleep, and Critical Care Journals. *Crit Care Med* 2020; 48(5): 623-33.
20. Levin A, Stevens PE, Bilous RW, et al. Kidney Disease: Improving Global Outcomes (KDIGO) CKD Work Group. KDIGO 2012 clinical practice guideline for the evaluation and management of chronic kidney disease. *Kidney international supplements* 2013; 3(1): 1-150.

21. Inker LA, Eneanya ND, Coresh J, et al. New creatinine-and cystatin C–based equations to estimate GFR without race. *New England Journal of Medicine* 2021; 385(19): 1737-49.
22. Ozrazgat-Baslanti T, Motaei A, Islam R, et al. Development and validation of computable phenotype to identify and characterize kidney health in adult hospitalized patients. *arXiv preprint arXiv:190303149* 2019.
23. Elixhauser A, Steiner C, Harris DR, Coffey RM. Comorbidity measures for use with administrative data. *Med Care* 1998; 36(1): 8-27.
24. Wald R, Waikar SS, Liangos O, Pereira BJ, Chertow GM, Jaber BL. Acute renal failure after endovascular vs open repair of abdominal aortic aneurysm. *J Vasc Surg* 2006; 43(3): 460-6; discussion 6.
25. Charlson ME, Pompei P, Ales KL, MacKenzie CR. A new method of classifying prognostic comorbidity in longitudinal studies: development and validation. *J Chronic Dis* 1987; 40(5): 373-83.
26. VHA V. National Drug File Reference Terminology (NDF-RT) Documentation. *US Department of Veterans Affairs* 2012.
27. Stottlemyer BA, Abebe KZ, Palevsky PM, et al. Expert Consensus on the Nephrotoxic Potential of 195 Medications in the Non-intensive Care Setting: A Modified Delphi Method. *Drug Saf* 2023; 46(7): 677-87.
28. Sundararajan M, Taly A, Yan Q. Axiomatic attribution for deep networks. International conference on machine learning; 2017: PMLR; 2017. p. 3319-28.
29. Lundberg SM, Lee S-I. A unified approach to interpreting model predictions. *Advances in neural information processing systems* 2017; 30.
30. Youden WJ. Index for rating diagnostic tests. *Cancer* 1950; 3(1): 32-5.
31. Christiadi D, Erlich J, Levy M, et al. The kinetic estimated glomerular filtration rate ratio predicts acute kidney injury. *Nephrology (Carlton)* 2021; 26(10): 782-9.
32. Edelstein CL. Biomarkers of acute kidney injury. *Adv Chronic Kidney Dis* 2008; 15(3): 222-34.
33. Selby NM, Pannu N. Opportunities in digital health and electronic health records for acute kidney injury care. *Curr Opin Crit Care* 2022; 28(6): 605-12.
34. Kashani KB, Koyner JL. Digital health utilities in acute kidney injury management. *Curr Opin Crit Care* 2023; 29(6): 542-50.
35. Yu X, Ji Y, Huang M, Feng Z. Machine learning for acute kidney injury: Changing the traditional disease prediction mode. *Frontiers in Medicine* 2023; 10: 1050255.
36. Caragata R, Wyssusek KH, Kruger P. Acute kidney injury following liver transplantation: a systematic review of published predictive models. *Anaesth Intensive Care* 2016; 44(2): 251-61.
37. Song X, Liu X, Liu F, Wang C. Comparison of machine learning and logistic regression models in predicting acute kidney injury: A systematic review and meta-analysis. *Int J Med Inform* 2021; 151: 104484.
38. Feng Y, Wang AY, Jun M, et al. Characterization of Risk Prediction Models for Acute Kidney Injury: A Systematic Review and Meta-analysis. *JAMA Network Open* 2023; 6(5): e2313359-e.
39. Uchino E, Sato N, Okuno Y. Artificial Intelligence in Predicting Kidney Function and Acute Kidney Injury. In: Lidströmer N, Ashrafian H, eds. Artificial Intelligence in Medicine. Cham: Springer International Publishing; 2020: 1-17.
40. Wainstein M, Flanagan E, Johnson DW, Shrapnel S. Systematic review of externally validated machine learning models for predicting acute kidney injury in general hospital patients. *Frontiers in Nephrology* 2023; 3.
41. Churpek MM, Carey KA, Edelson DP, et al. Internal and external validation of a machine learning risk score for acute kidney injury. *JAMA network open* 2020; 3(8): e2012892-e.

42. Song X, Yu AS, Kellum JA, et al. Cross-site transportability of an explainable artificial intelligence model for acute kidney injury prediction. *Nature communications* 2020; 11(1): 5668.
43. Kim K, Yang H, Yi J, et al. Real-time clinical decision support based on recurrent neural networks for in-hospital acute kidney injury: External validation and model interpretation. *Journal of Medical Internet Research* 2021; 23(4): e24120.
44. Mistry NS, Koyner JL. Artificial Intelligence in Acute Kidney Injury: From Static to Dynamic Models. *Adv Chronic Kidney Dis* 2021; 28(1): 74-82.
45. Bajaj T, Koyner JL. Artificial Intelligence in Acute Kidney Injury Prediction. *Advances in Chronic Kidney Disease* 2022; 29(5): 450-60.

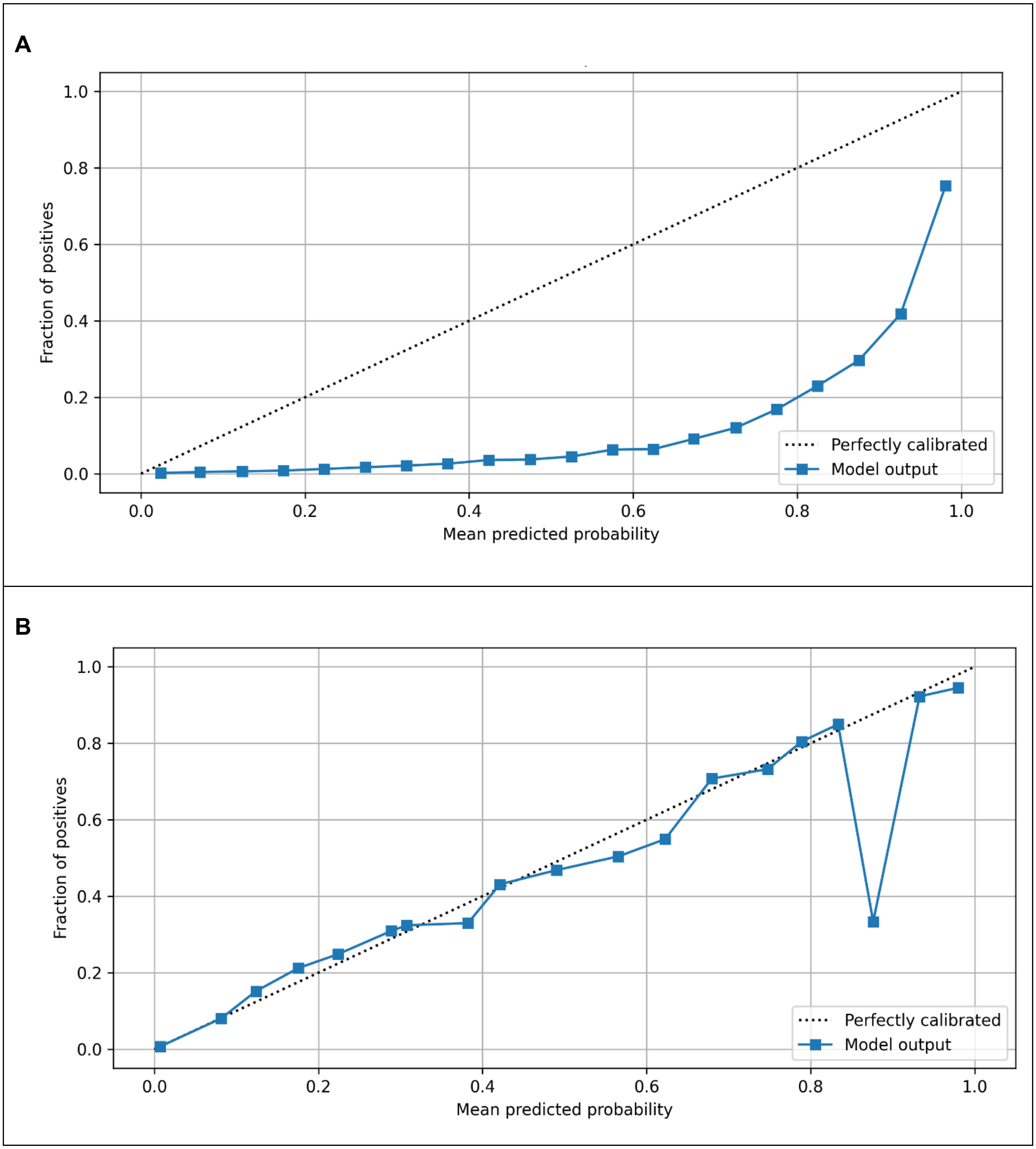

**Supplementary Figure 1.** Calibration plots for Model 3 using isotonic regression for 20 bins. A. Predictions before calibration, B. Predictions after calibration.

**Supplementary Table 1.** Summary of input features and preprocessing descriptions.

| Variable Name | Type | Data Source | Preprocessing |
|---|---|---|---|
| **Demographic Variables** | | | |
| Age | Continuous | Derived | Outlier adjustment [a], Feature scaling [c] |
| Gender | Binary | Raw | |
| Race | Nominal | Raw | Missing value imputation [b], One-hot encoding [d] |
| Ethnicity | Binary | Raw | Missing value imputation [b] |
| Language | Binary | Raw | Missing value imputation [b] |
| Insurance | Nominal | Raw | One-hot encoding [d] |
| **Admission Information** | | | |
| Admission source | Binary | Raw | |
| **Admission Comorbidities** | | | |
| Myocardial infarction | Binary | Derived | |
| Congestive heart failure | Binary | Derived | |
| Peripheral vascular disease | Binary | Derived | |
| Cerebrovascular disease | Binary | Derived | |
| Chronic pulmonary disease | Binary | Derived | |
| Metastatic carcinoma | Binary | Derived | |
| Cancer | Binary | Derived | |
| Liver disease | Binary | Derived | |
| Diabetes | Binary | Derived | |
| Hypertension | Binary | Derived | |
| Hypothyroidism | Binary | Derived | |
| Valvular disease | Binary | Derived | |
| Coagulopathy | Binary | Derived | |
| Obesity | Binary | Derived | |
| Weight loss | Binary | Derived | |
| Fluid/electrolyte disorders | Binary | Derived | |
| Chronic anemia | Binary | Derived | |
| Alcohol or drug abuse | Binary | Derived | |
| Depression | Binary | Derived | |
| Charlson comorbidity index | Continuous | Derived | Feature scaling [c] |
| Chronic kidney disease | Binary | Derived | |
| **Medications History** | | | |
| ACE Inhibitors | Binary | Derived | |
| Aminoglycosides | Binary | Derived | |
| Antiemetics | Binary | Derived | |
| Aspirin | Binary | Derived | |
| Beta Blockers | Binary | Derived | |
| Bicarbonates | Binary | Derived | |
| Corticosteroids | Binary | Derived | |
| Diuretics | Binary | Derived | |
| NSAIDS | Binary | Derived | |
| Vasopressors/Inotropes | Binary | Derived | |
| Statins | Binary | Derived | |
| Vancomycin | Binary | Derived | |
| Nephrotoxic | Binary | Derived | |
| Total number of medication types | Continuous | Derived | Feature scaling [c] |
| **Laboratory Results History** | | | |
| Hemoglobin | Continuous | Raw | Outlier adjustment [a], Missing value imputation [b], Feature scaling [c] |
| Leukocytes | Continuous | Raw | Outlier adjustment [a], Missing value imputation [b], Feature scaling [c] |
| Hematocrit | Continuous | Raw | Outlier adjustment [a], Missing value imputation [b], Feature scaling [c] |
| Erythrocyte mean corpuscular volume | Continuous | Raw | Outlier adjustment [a], Missing value imputation [b], Feature scaling [c] |
| Erythrocyte distribution width | Continuous | Raw | Outlier adjustment [a], Missing value imputation [b], Feature scaling [c] |
| Platelets | Continuous | Raw | Outlier adjustment [a], Missing value imputation [b], Feature scaling [c] |
| Glucose, serum | Continuous | Raw | Outlier adjustment [a], Missing value imputation [b], Feature scaling [c] |
| Urea nitrogen, serum | Continuous | Raw | Outlier adjustment [a], Missing value imputation [b], Feature scaling [c] |
| Creatinine, serum | Continuous | Raw | Outlier adjustment [a], Missing value imputation [b], Feature scaling [c] |
| Sodium, serum | Continuous | Raw | Outlier adjustment [a], Missing value imputation [b], Feature scaling [c] |
| Potassium, serum | Continuous | Raw | Outlier adjustment [a], Missing value imputation [b], Feature scaling [c] |
| Chloride, serum | Continuous | Raw | Outlier adjustment [a], Missing value imputation [b], Feature scaling [c] |
| Carbon dioxide, serum | Continuous | Raw | Outlier adjustment [a], Missing value imputation [b], Feature scaling [c] |
| Lactate, serum | Continuous | Raw | Outlier adjustment [a], Missing value imputation [b], Feature scaling [c] |
| Calcium, serum | Continuous | Raw | Outlier adjustment [a], Missing value imputation [b], Feature scaling [c] |
| Alanine aminotransferase, serum | Continuous | Raw | Outlier adjustment [a], Missing value imputation [b], Feature scaling [c] |
| Albumin, serum | Continuous | Raw | Outlier adjustment [a], Missing value imputation [b], Feature scaling [c] |
| Aspartate aminotransferase, serum | Continuous | Raw | Outlier adjustment [a], Missing value imputation [b], Feature scaling [c] |
| Bilirubin direct, serum | Continuous | Raw | Outlier adjustment [a], Missing value imputation [b], Feature scaling [c] |
| **Physiologic Variables** | | | |
| Mean arterial pressure | Continuous | Raw | Temporal processing [i], Time series creation [j], Feature scaling [c], Baseline time series extraction [k] |
| Heart rate | Continuous | Raw | Temporal processing [i], Time series creation [j], Feature scaling [c], Baseline time series extraction [k] |
| Respiration rate | Continuous | Raw | Temporal processing [i], Time series creation [j], Feature scaling [c], Baseline time series extraction [k] |
| Core temperature | Continuous | Raw | Temporal processing [i], Time series creation [j], Feature scaling [c], Baseline time series extraction [k] |
| **Laboratory Results** | | | |
| Sodium, serum | Continuous | Raw | Temporal processing [i], Time series creation [j], Feature scaling [c], Baseline time series extraction [k] |
| Potassium, serum | Continuous | Raw | Temporal processing [i], Time series creation [j], Feature scaling [c], Baseline time series extraction [k] |
| Carbon dioxide, serum | Continuous | Raw | Temporal processing [i], Time series creation [j], Feature scaling [c], Baseline time series extraction [k] |
| Chloride, serum | Continuous | Raw | Temporal processing [i], Time series creation [j], Feature scaling [c], Baseline time series extraction [k] |
| Glucose, serum | Continuous | Raw | Temporal processing [i], Time series creation [j], Feature scaling [c], Baseline time series extraction [k] |
| Calcium, serum | Continuous | Raw | Temporal processing [i], Time series creation [j], Feature scaling [c], Baseline time series extraction [k] |
| Urea nitrogen, serum | Continuous | Raw | Temporal processing [i], Time series creation [j], Feature scaling [c], Baseline time series extraction [k] |
| Hemoglobin | Continuous | Raw | Temporal processing [i], Time series creation [j], Feature scaling [c], Baseline time series extraction [k] |
| Leukocytes | Continuous | Raw | Temporal processing [i], Time series creation [j], Feature scaling [c], Baseline time series extraction [k] |
| Hematocrit | Continuous | Raw | Temporal processing [i], Time series creation [j], Feature scaling [c], Baseline time series extraction [k] |
| Platelets | Continuous | Raw | Temporal processing [i], Time series creation [j], Feature scaling [c], Baseline time series extraction [k] |
| Platelet mean volume | Continuous | Raw | Temporal processing [i], Time series creation [j], Feature scaling [c], Baseline time series extraction [k] |
| Erythrocyte mean corpuscular volume | Continuous | Raw | Temporal processing [i], Time series creation [j], Feature scaling [c], Baseline time series extraction [k] |
| Erythrocyte distribution width | Continuous | Raw | Temporal processing [i], Time series creation [j], Feature scaling [c], Baseline time series extraction [k] |
| Albumin, serum | Continuous | Raw | Temporal processing [i], Time series creation [j], Feature scaling [c], Baseline time series extraction [k] |
| Bilirubin total | Continuous | Raw | Temporal processing [i], Time series creation [j], Feature scaling [c], Baseline time series extraction [k] |
| **Miscellaneous** | | | |
| Nephrotoxic burden | Continuous | Derived | Temporal processing [i], Time series creation [j], Feature scaling [c], Baseline time series extraction [k] |
| Length of hospital stay | Continuous | Derived | Temporal processing [i], Time series creation [j], Feature scaling [c], Baseline time series extraction [k] |
| Kinetic estimated glomerular filtration | Continuous | Derived | Temporal processing [i], Time series creation [j], Feature scaling [c], Baseline time series extraction [k] |

[a] For continuous variables, values that fell in the top and bottom 1% of its distribution were considered outliers and capped to the respective values given at the 1st and 99th percentiles.

[b] Missing numerical values were replaced with the median from the development cohort, and missing nominal variables were assigned to a distinct “missing” category.
[c] Continuous variables were standardized to zero mean and unit variance.
[d] Nominal variables with less than 10 levels were represented as zero vectors of length equal to the number of levels, with level indicators equal to one.
[e] Using residency zip code, we linked to US Census data to calculate residing neighborhood characteristics and distance from hospital.
[f] Nominal variables with 10 levels or greater were transformed to a numeric integer identifier ranging from 0 to the number of unique levels minus one, where implicit variable representations were learned as part of the model training process.
[g] To preserve relative proximity, temporally recurring features such as month and day of admission were cyclically embedded as two separate features by sine and cosine-based transformation. For example, December (12) is near January (1), and Sunday (7) is near Monday (1).
[h] Medications were taken within one year timeframe prior to surgery using RxNorms data grouped into drug classes according to the US, Department of Veteran Affairs National Drug File Reference Terminology.
[i] Measurement values lying outside of expert-defined clinically normal value ranges for each variable were discarded. If two measurements existed at the same timestamp for a given patient, a random measurement was kept.
[j] For each surgical encounter, a time series was constructed by arranging measurements taken during hospitalization chronologically, summarizing the statistics for each day, performing linear interpolation in both directions and imputing the development median at every timestep for procedures lacking a single measurement of a particular variable.
[k] For baseline models, a set of five statistical features was extracted from each intraoperative time series. This set included the following features: minimum, maximum, mean, median, standard deviation.

**Supplementary Table 2.** Transition probability for AKI stage outcome within next 48 hours.

| | AKI stage within next 48 hours | | | | |
|---|---|---|---|---|---|
| | **UFH cohort** | | | | |
| **Current AKI stage** | No AKI | Stage 1 | Stage 2 | Stage 3 | Stage 3 + RRT |
| No AKI (n=836,245) | 809,602 (97%) | 23,918 (3%) | 1,901 (0.2%) | 793 (0.1%) | 31 (0.0%) |
| Stage 1 (n=41,488) | 8,037 (19%) | 30,532 (74%) | 2,447 (6%) | 445 (1%) | 27 (0.1%) |
| Stage 2 (n=10,520) | 495 (5%) | 1,493 (14%) | 7,725 (73%) | 773 (7%) | 34 (0.3%) |
| Stage 3 (n=7,295) | 195 (3%) | 148 (2%) | 516 (7%) | 6,223 (85%) | 213 (3%) |
| Stage 3 + RRT (n=1,849) | 4 (0.2%) | 1 (0.1%) | 1 (0.1%) | 4 (0.2%) | 1839 (99%) |
| | **UPMC cohort** | | | | |
| **Current AKI stage** | No AKI | Stage 1 | Stage 2 | Stage 3 | Stage 3 + RRT |
| No AKI (n=283,378) | 263,369 (93%) | 17,776 (6%) | 1,792 (1%) | 416 (0.2%) | 25 (0.01%) |
| Stage 1 (n=37,429) | 7,108 (19%) | 26,318 (70%) | 3,659 (10%) | 313 (1%) | 31 (0.1%) |
| Stage 2 (n=10,008) | 363 (4%) | 1,284 (13%) | 7,701 (77%) | 626 (6%) | 34 (0.3%) |
| Stage 3 (n=2,713) | 25 (1%) | 34 (1%) | 243 (9%) | 2,305 (85%) | 106 (4%) |
| Stage 3 + RRT (n=845) | 5 (1%) | 0 (0.0%) | 1 (0.1%) | 0 (0.0%) | 839 (99%) |